\documentclass[journal]{IEEEtran}
\usepackage{color}
\usepackage{graphicx}
\usepackage{amsmath}
\usepackage{array}
\usepackage{caption}
\usepackage{amsfonts}
\usepackage{amssymb}
\usepackage{epstopdf}
\begin{document}
 
\title{GAL: A Global-Attributes Assisted Labeling System for Outdoor Scenes}

\author{Yuzhuo~Ren, Chen~Chen, Shangwen~Li, and 
C.-C. Jay Kuo~\IEEEmembership{Fellow,~IEEE}
\thanks{The authors are with the Ming Hsieh Department of Electrical
Engineering, University of Southern California, Los Angeles, CA, 90089}
}
 
\maketitle
 
\begin{abstract}
An approach that extracts global attributes from outdoor images to
facilitate geometric layout labeling is investigated in this work. The
proposed Global-attributes Assisted Labeling (GAL) system exploits both
local features and global attributes. First, by following a classical
method, we use local features to provide initial labels for all
super-pixels. Then, we develop a set of techniques to extract global
attributes from 2D outdoor images. They include sky lines, ground lines,
vanishing lines, etc. Finally, we propose the GAL system that integrates
global attributes in the conditional random field (CRF) framework to
improve initial labels so as to offer a more robust labeling result. The
performance of the proposed GAL system is demonstrated and benchmarked with
several state-of-the-art algorithms against a popular outdoor scene layout
dataset.
\end{abstract}

\begin{IEEEkeywords}
Outdoor Layout Estimation, Semantic Labeling, Global Attribute Vector,
Convolutional Neural Network, 3D Reconstruction. 
\end{IEEEkeywords}
 
\IEEEpeerreviewmaketitle
 
\section{Introduction} 

\IEEEPARstart{A}{utomatic} 3D geometric labeling or layout reasoning
from a single scene image is one of the most important and challenging
problems in scene understanding. It offers the mid-level information for
other high-level scene understanding tasks such as 3D world
reconstruction\cite{choi2013understanding}, \cite{RefWorks:410},
\cite{hedau2009recovering}, \cite{RefWorks:390}, \cite{RefWorks:407},
\cite{park2015line}, depth map estimation \cite{RefWorks:397},
\cite{eigen2014depth}, \cite{ladicky2014pulling}, scene classification \cite{zou2016scene}, \cite{zang2015novel}, \cite{chen2016outdoor}
and content-based image retrieval \cite{gururaj2016effective}, \cite{johnson2015image}, \cite{paulin2015local}. 

Recovering the 3D structure from a single image using a local visual
pattern recognition approach was studied in early geometric labeling
research, including
\cite{RefWorks:410}, \cite{RefWorks:397}, \cite{RefWorks:393}, 
\cite{RefWorks:392}, \cite{RefWorks:406},
\cite{RefWorks:396}, \cite{RefWorks:401},
\cite{RefWorks:386}.  To give an example, Hoiem et al.
\cite{RefWorks:392} defined seven labels (i.e., sky, support, planar
left/right/center, porous and solid) and classified super-pixels to one
of these labels according to their local visual appearances. Features
such as color, position, texture pattern and segment shape were used to
describe local visual properties. Since the same surface (e.g., a
building facade) may take a different geometric role in different
images, the performance of all local-patch-based labeling methods is
limited. 

To improve the performance of local-patch-based methods, researchers
incorporated the global constraints or context rules in recent
years. Gupta et al. \cite{RefWorks:389}, \cite{RefWorks:395} proposed a
qualitative physical model for outdoor scenes by assuming that objects
are composed by blocks of volume and mass. If a scene image fits the
underlying 3D model, better surface layout estimation can be achieved.
However, their model is not generic enough to cover a wide range of
scenes. Liu et al. \cite{RefWorks:402}, and Pan et al. \cite{RefWorks:385}
focused on images containing building facades and developed multiple
global 3D context rules using their distinctive geometric cues such as
vanishing lines. Furthermore, recovering the 3D structure from depth
estimation algorithms was examined in \cite{eigen2014depth} and
\cite{ladicky2014pulling}. 

Recently, researchers applied the convolutional neural networks (CNN) to
the semantic segmentation task \cite{long2015fully},
\cite{badrinarayanan2015segnet}, \cite{zheng2015conditional},
\cite{farabet2013learning}, \cite{pinheiro2013recurrent},
\cite{im2016generating}, \cite{noh2015learning} and
\cite{dai2015instance}. Some improvement over traditional machine
learning based methods are observed for object-centric images and road
images. However, we have so far not yet seen a robust performance of the
CNN-based solution to semantic outdoor scene labeling. This is probably
due to the fact that it demands a large number of labeled scene images
to do the training and such a dataset is still not available. 

Being motivated by recent trends, we exploit both local and global
attributes for layout estimation and propose a Global-attributes
Assisted Labeling (GAL) system in this work. GAL uses local visual
pattern to provide initial labels for all pixels and extracts global
attributes such as sky lines, ground lines, horizon, vanishing lines,
etc. Then, it uses global attributes to improve initial labels.  Our
work contributes to this field in two folds. First, it provides a new
framework to address the challenging geometric layout labeling problem,
and this framework is supported by encouraging results. Second, as
compared with previous work, GAL can handle images of more diversified
contents using inference from global attributes. The performance of the
GAL system is benchmarked with several state-of-the-art algorithms
against a popular outdoor scene layout dataset, and significant
performance improvement is observed. 

The rest of this paper is organized as follows.  We first give a review
on previous work in Sec.~\ref{sec:GAL_relatedwork}. The GAL system is
described in Sec.~\ref{sec:overview}. Experimental results are shown in
Sec.~\ref{sec:result}, which includes analysis of several poor results.
Finally, concluding remarks are given in Sec.~\ref{sec:conclusion}. 

\section{Review of Previous Work}\label{sec:GAL_relatedwork}

\subsection{Local-Patch-based Labeling}

Hoiem et al. \cite{RefWorks:392} designed super-pixel level features and
used boosted decision tree classifiers to find the most likely label to
each super-pixel. The super-pixel segmentation has two limitations in the
geometric labeling problem. Fist, different regions with weak boundaries
subject to under-segmentation while texture regions subject to
over-segmentation. Second, since one segment is only annotated with a
single semantic label, if a wrong decision is made, the loss is huge.

One segmentation result cannot be perfect in geometric labeling as
pointed in \cite{RefWorks:392}. To overcome this problem, an algorithm
using the weighted sum of decisions from multiple segmentation results
was proposed in \cite{RefWorks:392}. Specifically, given an image, a
popular graph-based segmentation algorithm
\cite{felzenszwalb2004efficient} was applied using several different
parameter settings, leading to multiple super-pixels. Then, an algorithm was proposed to merge them into different segmentation numbers. The labeling accuracy can improve by considering the weighted sum of
decisions from different segments. 

The super-pixel learning method attempts to establish the relation
between local super-pixel appearance and the desired label. However, due
to the lack of global information and global physical constraints, the
super-pixel learning algorithm may not give meaningful results. 
 
\subsection{Blocks World Modeling}

To overcome the limitation of local-patch based labeling, researchers incorporated the global constraints or context rules in recent years. Gupta et al. \cite{RefWorks:389, RefWorks:395} proposed a qualitative physical model for outdoor scene images. In the inference process, each segment was fitted into one
of the eight block-view classes. Other constraints included geometric
constraints, contact constraints, intra-class and stability constraints
and depth constraints. The geometric constraints were obtained from the
initial labeling result in \cite{RefWorks:392}. The contact constraints
were employed to measure the agreement of geometric properties with
ground and sky contact points. The intra-class and stability constraints
were introduced to measure physical stability within a single block and
against other blocks, respectively. The depth constraints were used to
measure the agreement of projection of blocks in the 2D image plane with
the estimated depth ordering. Given a candidate block, $B_{i}$, its
associated geometric properties and its relationship with other blocks
was estimated by minimizing the cost function defined by the constrains.
By fitting blocks into a physical world, the global information can be
added into geometric labeling so that the local patch labeling error can
be reduced.  However, these algorithms suffer from the limited number of
block models, which fails to cover all possibilities in the real world.
In addition, if a segment is fitted to a wrong block model, its label
could be totally wrong. 
 
\subsection{Grammar-based Parsing and Mergence}

Liu et al. \cite{RefWorks:402} proposed a Bayesian framework and five
merge rules to merge super-pixels in a bottom-up fashion for geometric
labeling of urban scenes. The algorithm found the straight lines,
estimated the vanishing points in the image, and partitioned the image
into super-pixels. The inference was done based on Composite Cluster
Sampling (CCS). The initial covering was obtained using the K-means
algorithm. Then, at each iteration, it made proposals based on five
grammar rules: layering (rule 1), siding (rule 2), supporting (rule 3),
affinity (rule 4) and mesh (rule 5). The five grammar rules are used to maximize the posterior probability in Bayesian formulation. The layering rule includes the focal length and camera height and describes the connection between the scene node and other super-pixels. Siding rule describes the spatial connection of two super-pixels and the contact line. Supporting rule states one super-pixel is supporting another if their surface normals are orthogonal. The affinity rule states that two super-pixels are likely to belong to the same surface if they have similar appearance. Mesh rule statues that multiple super-pixels are arranged in a mesh structure described by two orthogonal vanishing points. The Bayesian inference consisted of two stages. In the first stage, proposals were made using rule 4 and rule 5. In the second stage, proposals were made based on all five grammar rules. Since this algorithm heavily relies on the Manhattan world assumption as well as accurate vanishing point detection results, it cannot handle most nature scenes well. 
 
\subsection{3D Building Layout Inference}

Pan et al. \cite{RefWorks:385} focused on images containing building
facades.  Given an urban scene image, they first detected a set of
distinctive facade planes and estimated their 3D orientations and
locations. Being different from previous methods that provided coarse
orientation labels or qualitative block approximations, their algorithm
reconstructed building facades in the 3D space quantitatively using a
set of planes mutually related via 3D geometric constraints. Each plane
was characterized by a continuous orientation vector and a depth
distribution, and an optimal solution was searched through inter-planar
interaction. It inferred the optimal 3D layout of building facades by
maximizing a defined objective function. The data term was the product
of two scores, indicating the image feature compatibility and the
geometric compatibility, respectively. The former measured the agreement
between the 2D location of a facade plane and image features while the
latter measured the probability of a ground contact line position. The
smoothness term included the convex-corner constraint, the occlusion
constraint and the alignment constraint. By exploiting quantitative
plane-based geometric reasoning, this solution is more expressive and
informative than other methods.  However, it does not provide suitable
models for the labeling of the ground, sky, porous and solid classes in
general scenes.

\begin{figure*}[ht] 
\begin{center}
\includegraphics[width=1.0\textwidth]{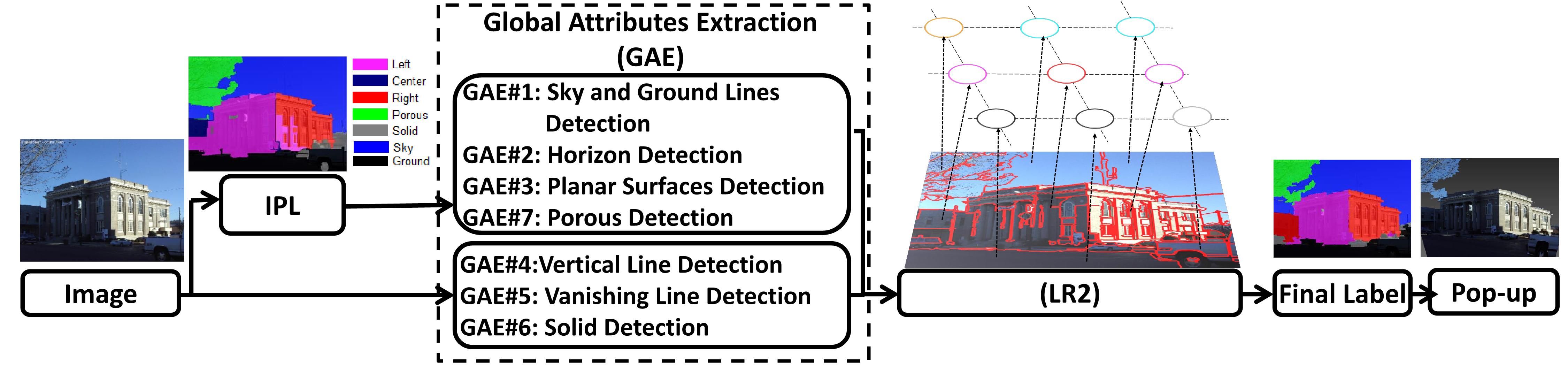}
\caption{The flow chart of the proposed GAL system.}
\label{fig:framework}
\end{center}
\end{figure*}
 	
\section{Proposed GAL System}\label{sec:overview}

\subsection{System Overview}

The flow chart of the GAL system is given in \figurename{\ref{fig:framework}}.
\begin{itemize}
\itemsep -0.3ex
\item \begin{flushleft}Stage 1: Initial Pixel Labeling (IPL);\end{flushleft}
\item \begin{flushleft}Stage 2: Global Attributes Extraction (GAE);\end{flushleft}
\item \begin{flushleft}Stage 3: Layout Reasoning and Label Refinement (LR2).\end{flushleft}
\end{itemize}

For a given outdoor scene image, we can obtain initial pixel labeling
results using pixel-wise labeling method in the first
stage. Here, we trained seven class labels use SegNet architecture
\cite{badrinarayanan2015segnet}. The reason that we use SegNet over other CNN based segmentation approaches\cite{long2015fully,zheng2015conditional} because of the balance of segmentation accuracy and efficiency\cite{badrinarayanan2015segnet}.
The labeling performance of the IPL stage is however not satisfactory due to the lack of global
scene information. To address this issue, we pose the following seven
questions for each scene image and would like to answer them based on
all possible visual cues (e.g., color, edge contour, defocus degree,
etc.) in the second stage:
\begin{enumerate}
	\itemsep -0.3ex
	\item Is there sky in the image? If yes, where?
	\item Is there ground in the image? If yes, where?
	\item Does the image contain a horizon? If yes, where?
	\item Are there planar surfaces in the image? If yes,
	where and what are their orientations?
	\item Is there any building in the image? If yes, where and
	what is its orientation?
	\item Is there solid in the image? If yes, where is it?
	\item Is there porous in the image? if yes, where is it?
\end{enumerate}

The answers to the first part of each question lead to a 7D global
attribute vector (GAV) with binary values (YES or NO), where we set
``YES" and ``NO" to ``1" and ``0", respectively. If the value for an
entry is ``1", we need to provide more detailed description for the
corresponding global attribute. The knowledge of the GAV is helpful in
providing a robust labeling result. Based on extracted global
attributes, we conduct layout reasoning and label refinement in the
third stage. Layout reasoning can be greatly simplified based on global
attributes. Then, the label of each pixel can be either confirmed or
adjusted based on inference. The design and extraction of global
attributes in the GAE stage and the layout reasoning and label
refinement in the LR2 stage are two novel contributions. They will be
elaborated in Sec. \ref{sec:GAE} and Sec. \ref{sec:LR2}, respectively.

\subsection{Initial Pixel Labeling (IPL)}\label{sec:IRL}

The method proposed by Hoiem in \cite{RefWorks:393} offers an excellent
candidate in the first stage to provide initial pixel-level labels of
seven geometric classes (namely; sky, support, planar left/right/center,
porous and solid). This method extracts color, texture and location
features from super-pixels, and uses a learning-based boosted decision
tree classifier.  We also tried the CNN approach \cite{long2015fully},
\cite{badrinarayanan2015segnet},
\cite{zheng2015conditional} to initialize pixel labels. However, due to
lacking enough training data, the CNN solution provides results much
worse than those in \cite{RefWorks:393}. 

\begin{figure}[t]
\begin{center}
\includegraphics[width=1\linewidth]{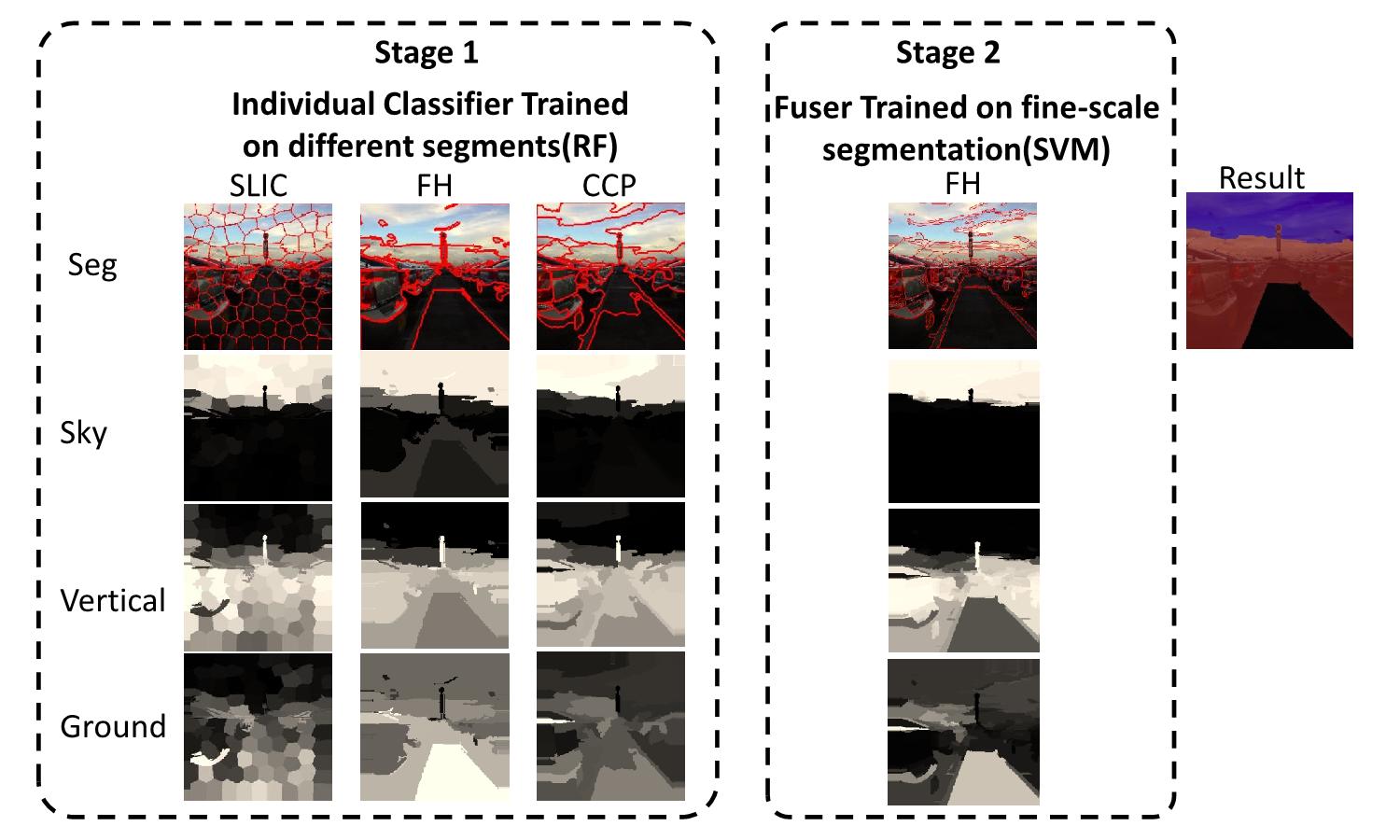}
\end{center}
\caption{Fusion of 3-class labeling algorithms. Stage 1: Three
individual random forest classifiers trained by segments from SLIC \cite{achanta2012slic}, FH \cite{felzenszwalb2004efficient}
and CCP \cite{fu2015robust}, respectively, where the gray images show the probability output
of each individual classifier under different segmentation methods and
different geometric classes. Stage 2: The three probability outputs from
Stage 1 are cascaded into one long feature vector for each intersected
segmentation unit and an SVM classifier is trained to get the final
decision.}\label{fig:3class}
\end{figure}

To enhance the prediction accuracy for sky and support in
\cite{RefWorks:393}, we develop a 3-class labeling scheme that
classifies pixels to three major classes; namely, ``support",
``vertical" and ``sky", where planar left/right/center, porous and solid
are merged into one ``vertical" mega-class. This 3-class classifier is
achieved by integrating segmentation results from SLIC
\cite{achanta2012slic}, FH \cite{felzenszwalb2004efficient}, and CCP
\cite{fu2015robust} with a random forest classifier
\cite{liaw2002classification} in a two-stage classification system.
\figurename{\ref{fig:3class}} shows the proposed two-stage system. In the
first stage, we train individual classifiers for ``support",``vertical"
and ``sky" and get their probability maps using the SLIC, FH and CCP
segmentations. In the second stage, since the segmentation units are not
the same, we transfer all segments into smaller segmentation units using
the FH method to obtain the fine-scale segmentation boundaries. Then, we
fuse the probability output from the first stage to get the final
decision. 

\begin{figure}[t]
\begin{center}
\includegraphics[width=1.0\linewidth]{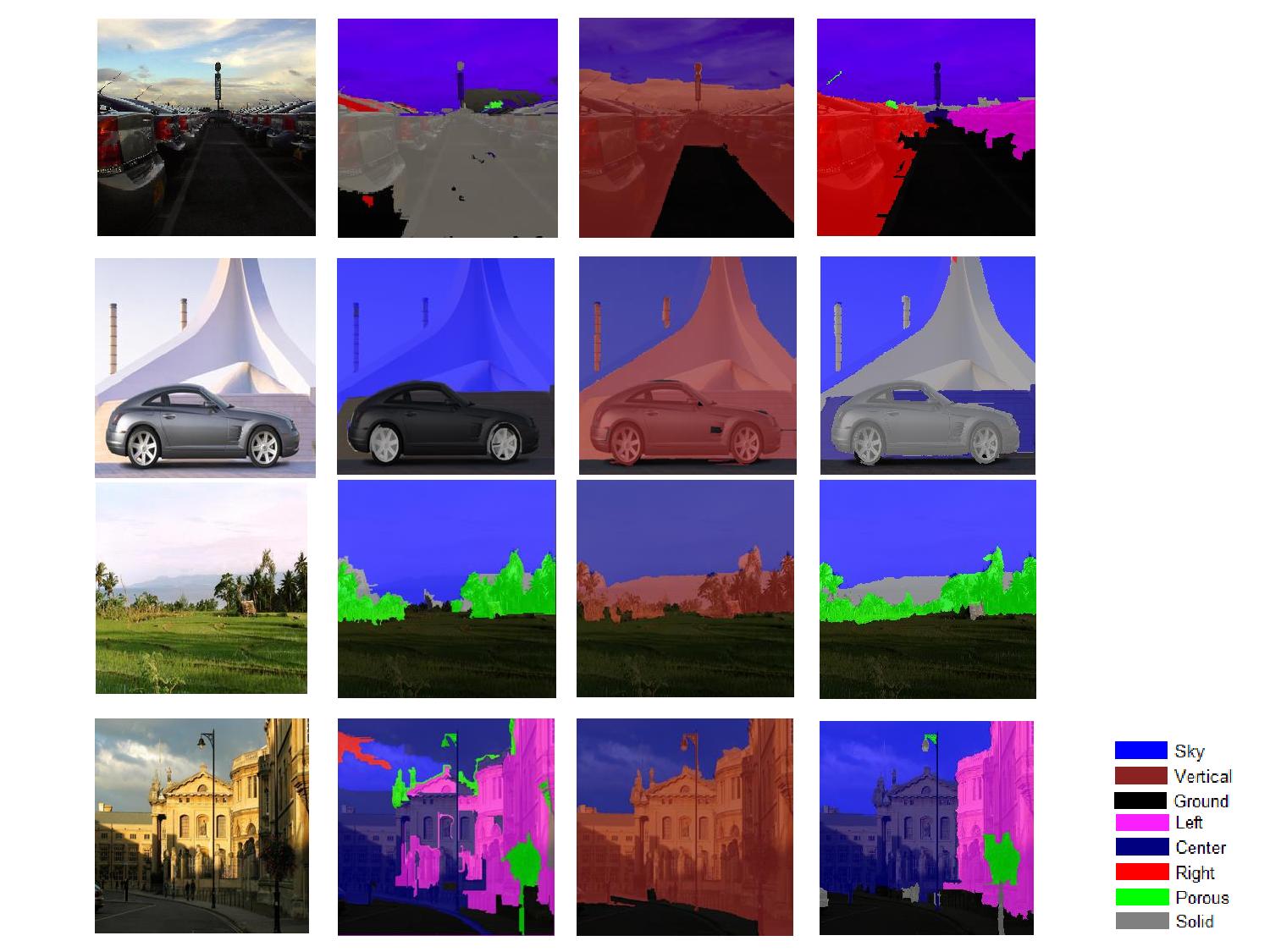}
\end{center}
\caption{Comparison of initial pixel labeling results (from left to
right): the original image, the 7-class labeling result from
\cite{RefWorks:393}, the proposed 3-class labeling result, the ground
truth 7-class labeling.  Our scheme offers better ``support" and ``sky"
labels.} \label{fig:3class_compare}
\end{figure}

The reason of combining different segmentation schemes is that 
different segmentation methods capture different levels of
information. For example, the SLIC method \cite{achanta2012slic}
provides more segments in the ``support" region than the FH and CCP
methods in the first stage. Thus, the SLIC segmentation has better
location information and a higher chance to get the correct result for
``support". The performance of our 3-class labeling scheme is 88.7\%,
which is better than that of \cite{RefWorks:393} by 2\%. Several visual
comparisons are shown in \figurename{\ref{fig:3class_compare}}. We see
that our 3-class labeling system can segment the low contrast ``support"
and ``sky" well. After the 3-class labeling, initial labels of the five
classes inside the vertical region come directly from \cite{RefWorks:393}. 

The accuracy of the IPL stage is highly impacted by three factors: 1) a
small number of training samples, 2) the weak discriminant power of
local features, and 3) lacking of a global scene structure. They are
common challenges encountered by all machine learning methods relying on
local features with a discriminative model. To overcome these
challenges, we design global attributes vectors and integrate them into
a graphical model, which is elaborated in the next subsection. 

\subsection{Global Attributes Extraction (GAE)}\label{sec:GAE}

In the second GAE stage, we attempt to fill out the 7D binary-valued GAV
and find the related information associated with an existing element.
The 7 global attributes are: 1) the sky/ground line, 2) the horizon, 3)
the planar surface, 4) the vertical line, 5) the vanishing line, 6) the
solid object, and 7) the porous material. Take image $I$ with dimension
$H \times W \times 3$ as an example. In the GAE stage, its 7D GAV will
be extracted to generate 7 probability maps denoted by $P_{k}$, $k=1,
\cdots, 7$, where the dimension of $P_{k}$ is $H \times W$ and $k$
represents one of the 7 global attributes.  Furthermore, we use
$P_{k}(s_{i},l_{j})$ to denote the probability for segment $s_{i}$
to be labeled as $l_{j}$, where $j$ denotes one of the 7 classes
(support, left, center, right, porous, solid and sky) based on global
attribute $k$. 

\begin{figure}[t]
\begin{center}
\includegraphics[width=1\linewidth]{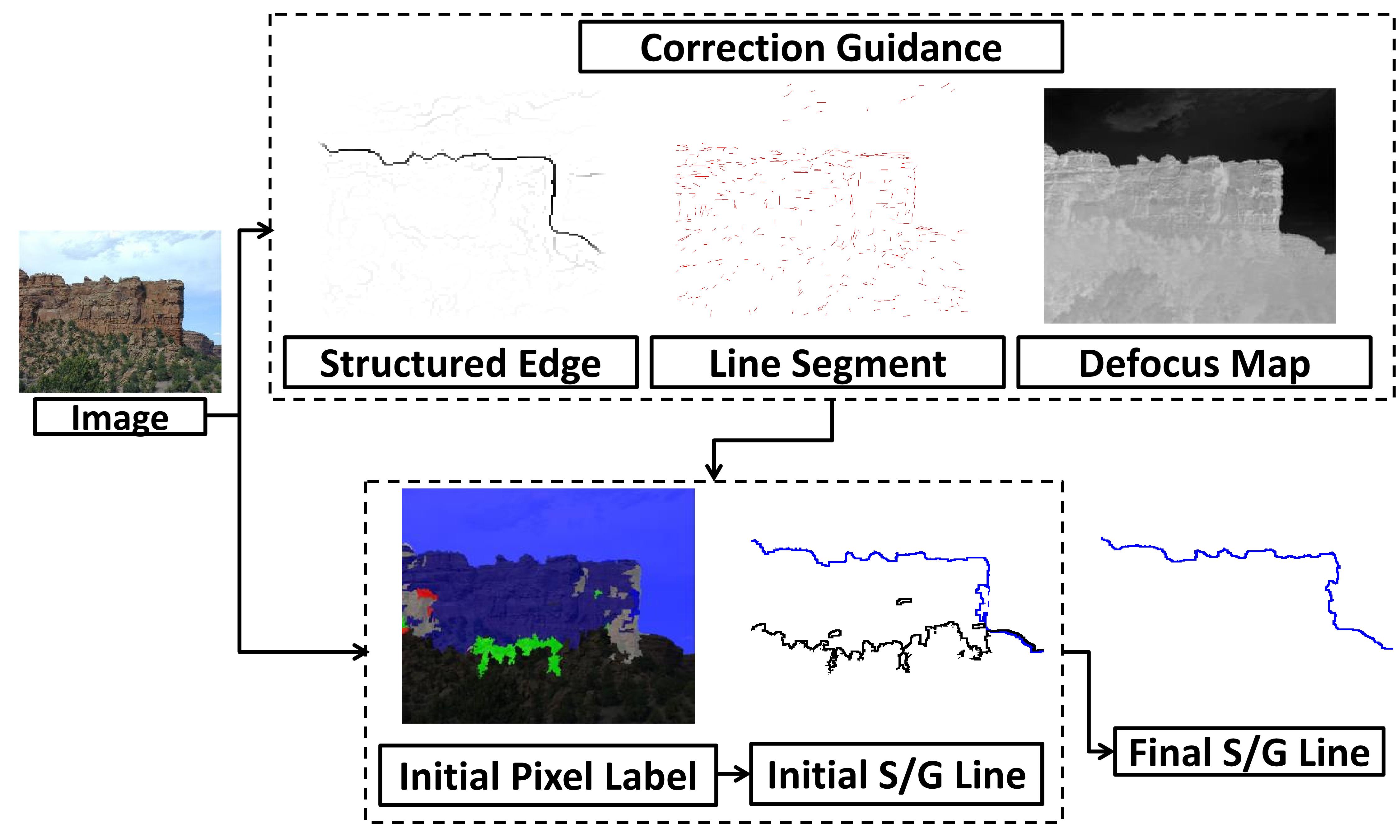}
\caption{The process of validating the existence of the sky/ground lines 
and their location inference.}\label{fig:sky_ground_existence}
\end{center}
\end{figure}

{\bf Sky and Ground Lines Detection.} Sky and ground regions are important
ingredients of the geometrical layout of both natural and urban scenes.
To infer their existence and correct locations is critical to the task
of scene understanding. We develop a robust procedure to achieve this
goal as illustrated in \figurename{\ref{fig:sky_ground_existence}}.
Based on initial pixel labels obtained in the first stage, we obtain
initial sky and ground lines, which may not be correct due to erroneous
initial labels.

To finetune initial sky and ground lines, we exploit the following 
three cues from the input scene image for sky and ground line validation:
\begin{itemize}
\item the line segment map denoted by $P_{LS}$ \cite{von2008lsd}, where
$P_{LS}=1$ for line pixels, $P_{LS}=0$ for non-line pixels;
\item the probability edge map of structured edges denoted by $P_{SE}$, 
\cite{dollar2014fast}, \cite{zitnick2014edge}, \cite{dollar2013structured}; and
\item the probability edge map of the defocus map denoted by $P_{DF}$
\cite{ZhuoAndSim2011}.
\end{itemize}
The final probability is defined as 
\begin{equation} 
P_{\text{sky/ground line}}(s_{i},l_{\text{sky/ground}})= P_{LS} \times P_{SE} \times P_{DF}.
\end{equation}
An example is given in \figurename{\ref{fig:sky_ground_existence}},
where all three maps have higher probability for the sky line but
lower probability scores for the ground line. As a result, the erroneous
ground line is removed. 


After obtaining the sky and ground lines, we check whether there is any
vertical region above the sky line or below the ground line using the
3-class labeling scheme, where the vertical region is either solid or
porous. The new 3-class labeling scheme can capture small vertical
regions well and, after that, we will zoom into each vertical region to
refine its subclass label.

{\bf Horizon Detection.} When the ground plane and the sky plane meet,
we see a horizon. This occurs in ocean scenes or scenes with a flat
ground. The horizon location helps reason the ground plane, vertical
plane and the sky plane. For example, the ground should not be above the
horizon while the sky should not be below the horizon. Generally
speaking, 3D layout estimation accuracy can be significantly enhanced if
the ground truth horizon is available for layout
reasoning\cite{RefWorks:393}. Research on horizon estimation from
building images was done before, {\em e.g.}, \cite{RefWorks:405},
\cite{RefWorks:404}. That is, it can be inferred by connecting
horizontal vanishing points. However, the same technique does not apply
to natural scene images where the vanishing point information is
lacking. 

\begin{figure}
\begin{center}
\includegraphics[width=1.0\linewidth]{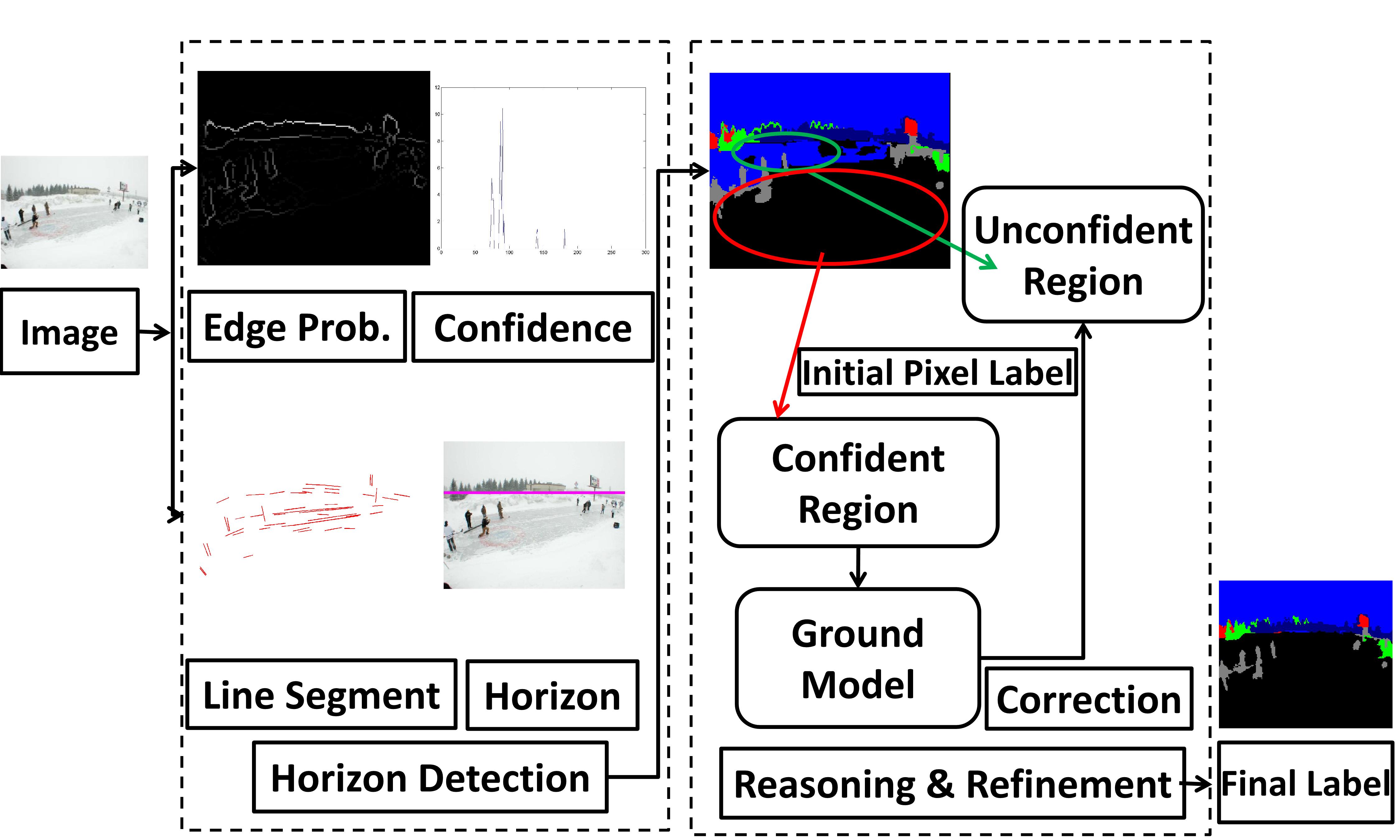}
\end{center}
\caption{Horizon detection and its application to layout reasoning and
label refinement.} \label{fig:horizon}
\end{figure}

In our implementation, we use two different methods to estimate the
horizon in two different outdoor scenes. For images containing buildings
as evidenced by strong vertical line segments, their horizon can be
estimated by fitting the horizontal vanishing points\cite{RefWorks:404}.
For natural scene images that do not have obvious vanishing points, we
propose a horizon estimation algorithm as shown in \figurename{\ref{fig:horizon}}. First, we extract multiple horizontal line segments
from the input image using the LSD algorithm \cite{von2008lsd}. Besides,
we obtain the edge probability map based on
\cite{dollar2014fast}, \cite{zitnick2014edge}, \cite{dollar2013structured} and use it as well as a
location prior to assign a probability to each pixel in the line
segments. Then, we build a histogram to indicate the probability of the
horizon location. Finally, we will select the most likely horizonal line
to be the horizon. 

After detecting the horizon, we perform layout reasoning to determine
the sky and support regions. One illustrative example is given in
\figurename{\ref{fig:horizon}}. We can divide the initial labeled
segments into two regions. If a segment above (or below) the horizon is
labeled as sky (or support), it belongs to the confident region.  On the
other hand, if a segment above (or below) the horizon is labeled as
support (or sky), it belongs to the unconfident region. The green
circled region in \figurename{\ref{fig:horizon}} is labeled as sky due to
its white color. Thus, it lies in the unconfident region. There exists a
conflict between the local and global decisions. To resolve the
conflict, we use the Gaussian Mixture Model (GMM) to represent the color
distribution in the confident region above and below the horizon.  Then,
we conclude that the white color under the horizon can actually be the
support (or ground) so that its label can be corrected accordingly.  We
use $P_{\text{horizon}}(s_{i},l_{j})$ to denote the probability output from the
GMM. 

{\bf Planar Surfaces Detection.} An important by-product of sky/ground
line localization is the determination of planar surface orientation in
the vertical region. This is feasible since the shapes of sky and ground
lines provide useful cues for planar surface orientation inference in
natural or urban scenes. To be more specific, we check the trapezoidal
shape fitting scheme (including triangles and rectangles as special
cases) for the vertical region, where the top and the bottom of the
trapezoidal shape are bounded by the sky and ground lines while its left
and right are bounded by two parallel vertical lines or extends to the
image boundary. We set $P_{\text{planar surface}}(s_{i},l_{j})=1$, where
$j\in{\text{(left,center,right)}}$, if the corresponding surface orientation is
detected. Otherwise, we set $P_{\text{planar surface}}(s_{i},l_{j})=0$. 

\begin{figure}
\begin{center}
\includegraphics[width=1.0\linewidth]{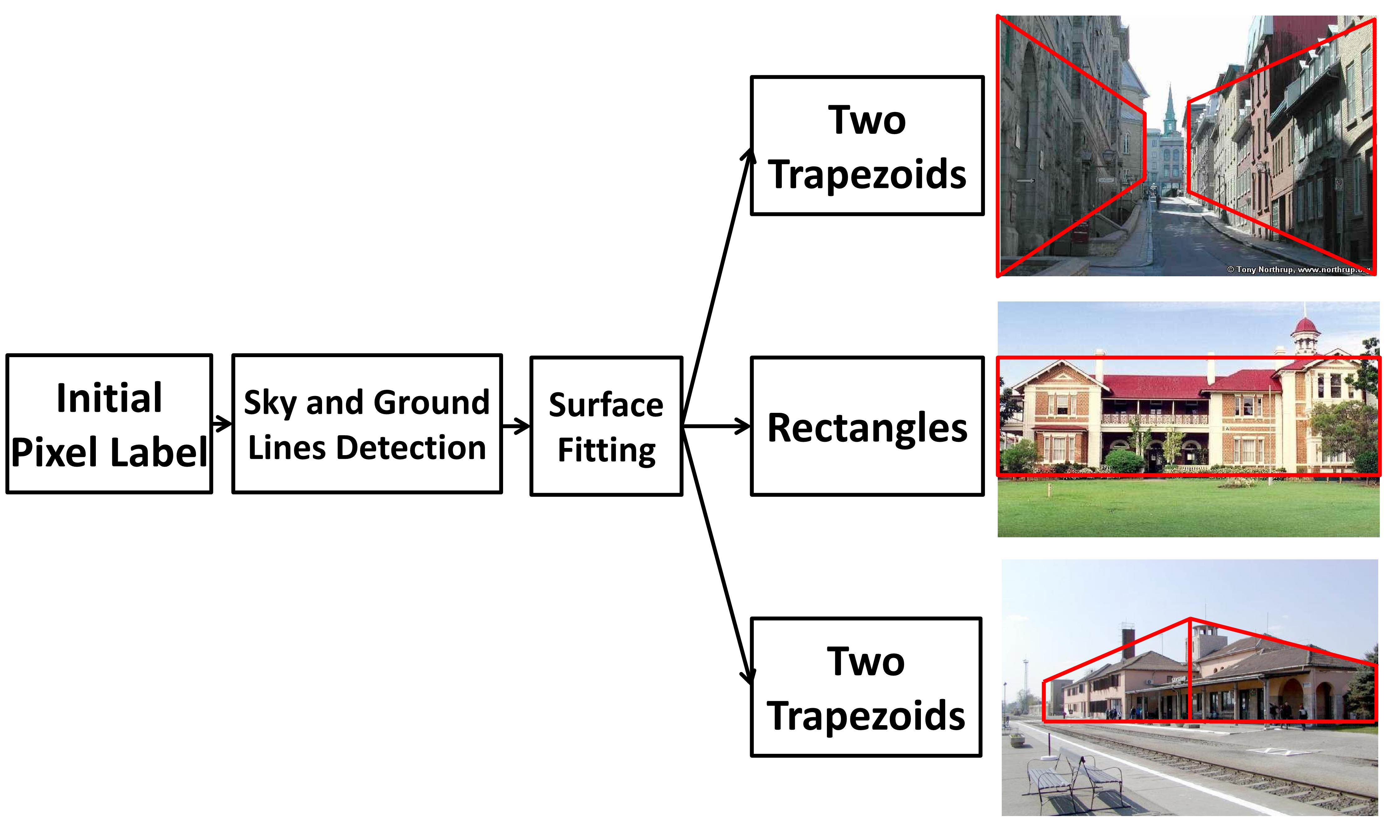}
\end{center}
\caption{Trapezoidal shape fitting with the sky line in the top
and the ground line in the bottom.}\label{fig:sky_ground_shape}
\end{figure}

Three trapezoidal region shape fitting examples are shown in
\figurename{\ref{fig:sky_ground_shape}}. Clearly, different fitting
shapes indicate different planar surface orientations. For example, two
trapezoidal regions with narrow farther sides indicate an alley scene as
shown in the top example. A rectangle shape indicates a front shot of a
building as shown in the middle example. The two trapezoidal regions with
one common long near side indicates two building facades with different
orientations as given in the bottom example. Thus, the shapes of sky and
ground lines offer important cues to planar surface orientations.

\begin{figure}
\begin{center}
\includegraphics[width=1.0\linewidth]{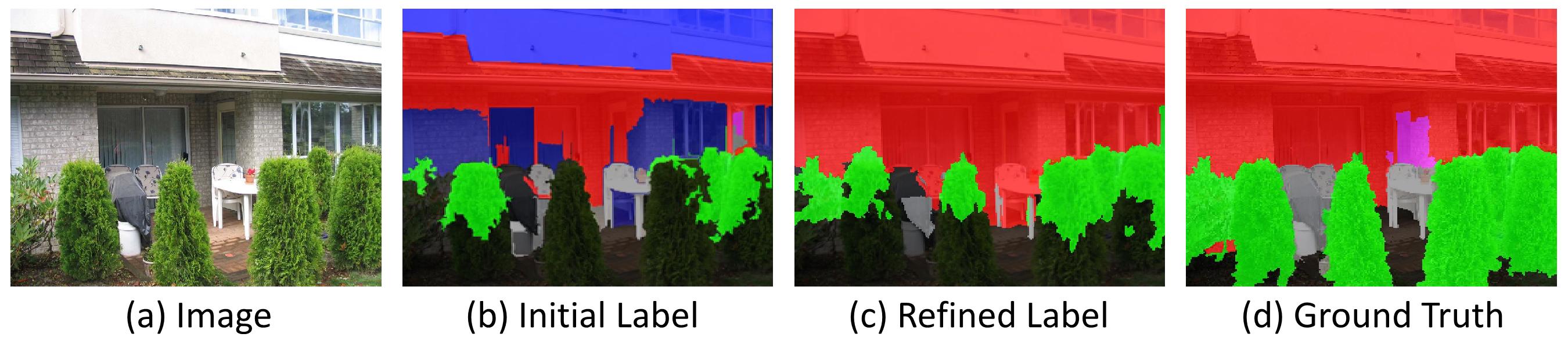}
\end{center}
\caption{An example of vertical line detection and correction.}\label{fig:VL}
\end{figure}

{\bf Vertical Line Detection.} A group of parallel vertical line
segments provides a strong indicator of a building structure in scene
images. It offers a valuable cue in correcting wrongly labeled regions
in building scene images.  The probability of surface orientation from
the vertical line is denoted by $P_{\text{verticall}}(s_{i},l_{j})$, where
$j\in{\text{(left, center, right)}}$. In our implementation, we use the
vertical line percentage in a region as the attribute to generate the
probability map for the building region. An example of using vertical
lines to correct wrongly labeled regions is illustrated in \figurename{\ref{fig:VL}}. The top region of the building is wrongly labeled as
``sky" because of the strong location and color cues in the initial
labeling result. However, the same region has a strong vertical line
structure. Since the ``sky" region should not have this structure, we
can correct its wrong label. 

\begin{figure}
\begin{center}
\includegraphics[width=1.0\linewidth]{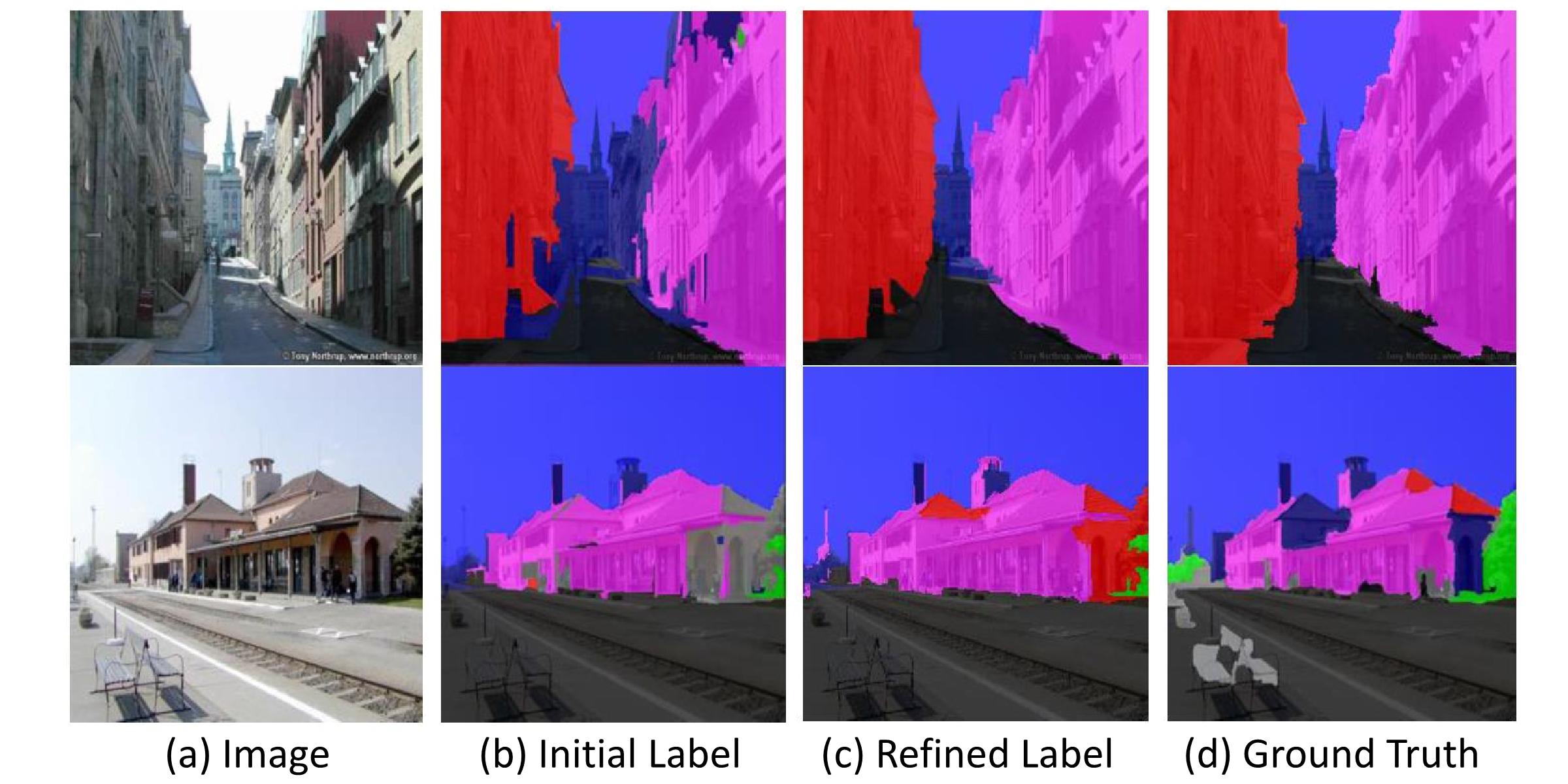}
\end{center}
\caption{Examples of surface orientation refinement using the shape 
and the vanishing line cues.} \label{fig:surface}
\end{figure}

{\bf Vanishing Line Detection.} A group of vanishing lines offers
another global attribute to indicate the surface orientation. In our
implementation, we first use the vertical line detection technique to
obtain a rough estimate of the left and right boundaries of a building
region and then obtain its top and bottom boundaries using the sky line
and the ground line. After that, we apply the vanishing line detection
algorithm in \cite{hedau2009recovering} and use the obtained vanishing
line result to adjust the orientation of a building facade. We set
$P_{\text{vanishingl}}(s_{i},l_{j})=1$, where $j\in{\text{(left,center,right)}}$, if
the corresponding surface orientation is detected. Otherwise, we set
$P_{\text{vanishingl}}(s_{i},l_{j})=0$. Note that the surface orientation of a
planar surface can be obtained from its shape and/or vanishing lines. If
it is not a building, we cannot observe vanishing lines and the shape is
the only cue. If it is a building we can observe both its shape and
vanishing lines. The two cues are consistent with each other based on
our experience. Two examples of using the shape and vanishing lines of a
building region to correct erroneous initial labels are given in
\figurename{\ref{fig:surface}}. The initial surface orientation
provided by the IPL stage contains a large amount of errors due to the
lack of global view. They are corrected using the global attributes. 

{\bf Solid Detection.} The non-planar solid class is typically composed
by foreground objects (such as people, car, etc.) rather than the
background scene. The bottom of an object either touches the ground or
extends to the image bottom boundary. For example, objects (say,
pedestrians and cars) may stand on the ground in front of the building
facade in an urban scene while objects are surrounded by the ground
plane in a typical natural scene. Object detection is an active research
field by itself. Actually, the object size has an influence on the scene
content, i.e. whether it is an object-centric or scene-centric image.  The
object size is big in an object-centric image. It occupies a large
portion of the image, and most of the background is occluded. The scene
layout problem is of less interest since the focus is on the object
rather than the background. In contrast, there is no dominant object in
a scene-centric image and the scene layout problem is more significant.
On one hand, it is better to treat the object detection problem
independently from scene analysis. On the other hand, a scene image may
still contain some objects while the background scene is occluded by
them. For examples, the existence of objects may occlude some part of
sky and ground lines, thus increasing the complexity of scene layout estimation. In order to simplify the scene understanding problem,
it is desired to identify and remove objects first. 

\begin{figure}[htbp]
\begin{center}
\includegraphics[width=1.0\linewidth]{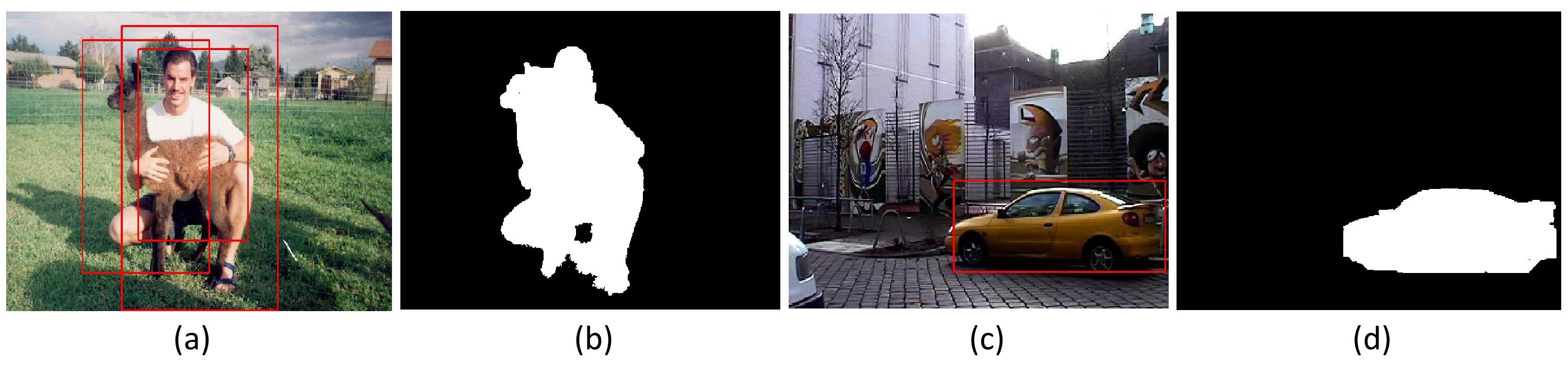}
\end{center}
\caption{Examples of obtaining the object mask using the person and the
car object detectors and the grab cut segmentation method: (a) the
bounding boxes result of the person detector, (b) the person
segmentation result, (c) the bounding box result of the car detector,
and (d) the car segmentation result.}\label{fig:object}
\end{figure}

In our implementation, we apply two object detectors (namely, the person
detector and the car detector) from \cite{voc-release5}
\cite{felzenszwalb2010object}. We first obtain the bounding boxes of detected
objects and, then, adopt the grab cut method \cite{rother2004grabcut} to
get their exact contours. The grab cut method uses a binary value to
indicate whether a pixel belong to object or not. Mathematically, we
have $P_{\text{solid}}(s_{i},l_{\text{solid}})=1$ for the location where solid is
detected, and $P_{\text{solid}}=0$, otherwise. Two examples are shown in
\figurename{\ref{fig:object}}. To achieve a better layout estimation,
detected and segmented objects are removed from the scene image to allow
more reliable sky and ground line detection. 

\begin{figure}[htbp]
\begin{center}
\includegraphics[width=1.0\linewidth]{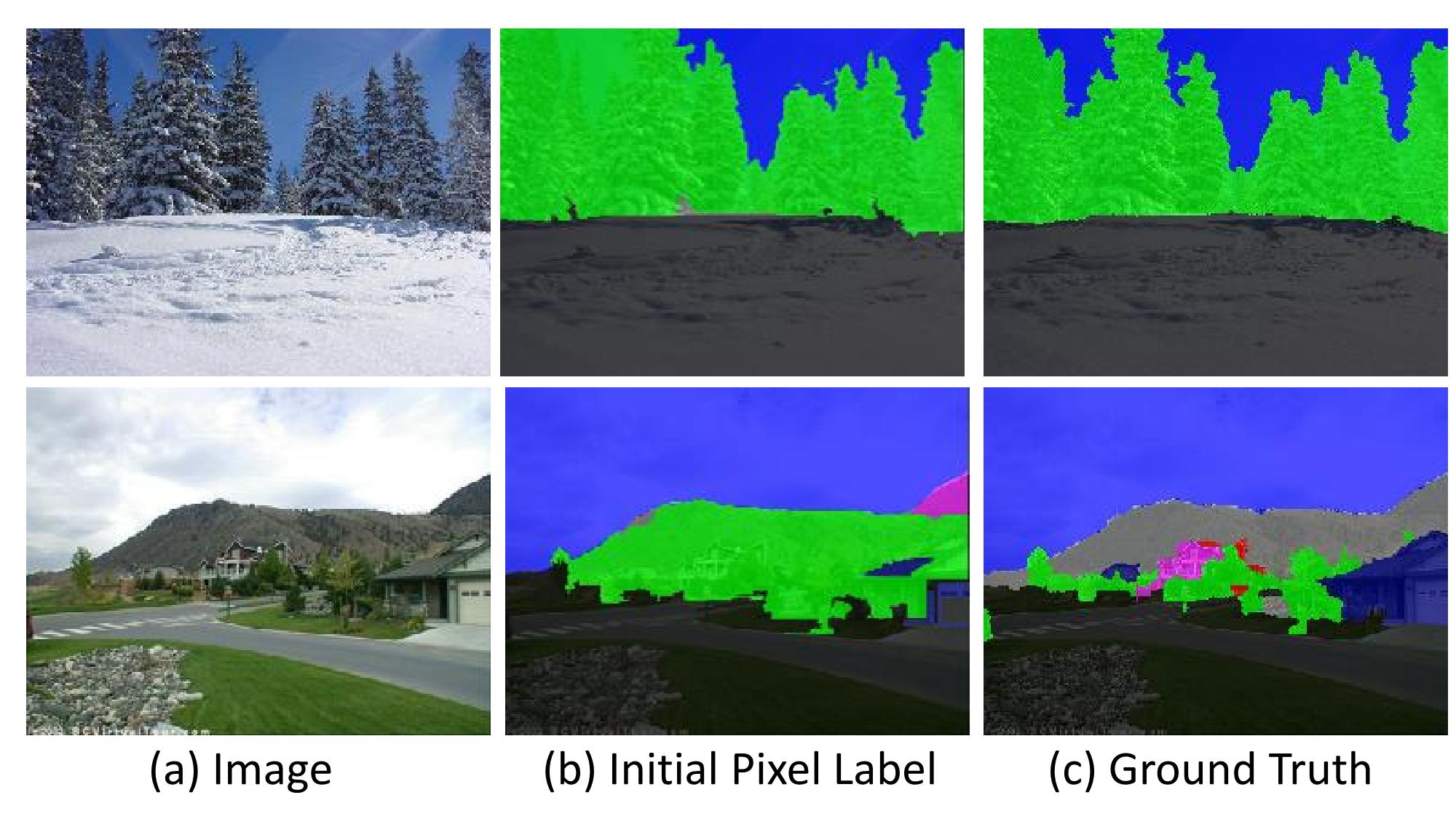}
\end{center}
\caption{Two examples of initially labeled porous regions where the top
(trees) is correctly labeled while the bottom (mountain) is wrongly
labeled.}\label{fig:porous}
\end{figure}

{\bf Porous Detection.} As shown in the bottom subfigure of
\figurename{\ref{fig:porous}}, the mountain region, which belongs to the
solid class, is labeled as porous by mistake. This is due to the fact
that the super-pixel segmentation method often merges the porous region
with its background. Thus, it is difficult to split them in the
classification stage. To overcome this difficulty, we add the contour
randomness feature from the structured edge result
\cite{dollar2014fast}, \cite{zitnick2014edge},
\cite{dollar2013structured} to separate the porous and the solid
regions. Note that $P_{\text{porous}}(s_{i},l_{\text{porous}})$ is proportional to
contour randomness. By comparing the two examples in
\figurename{\ref{fig:porous}}, we see that there exist irregular
contours inside the true porous region (trees) but regular contours
inside the solid region (mountain).  In our implementation, we double
check regions labeled by solid or porous initially and use the contour
smoothness/randomness to separate solid/porous regions. 

\subsection{Layout Reasoning and Label Refinement (LR2)}\label{sec:LR2}

\begin{figure}[htbp]
\begin{center}
\includegraphics[width=1\linewidth]{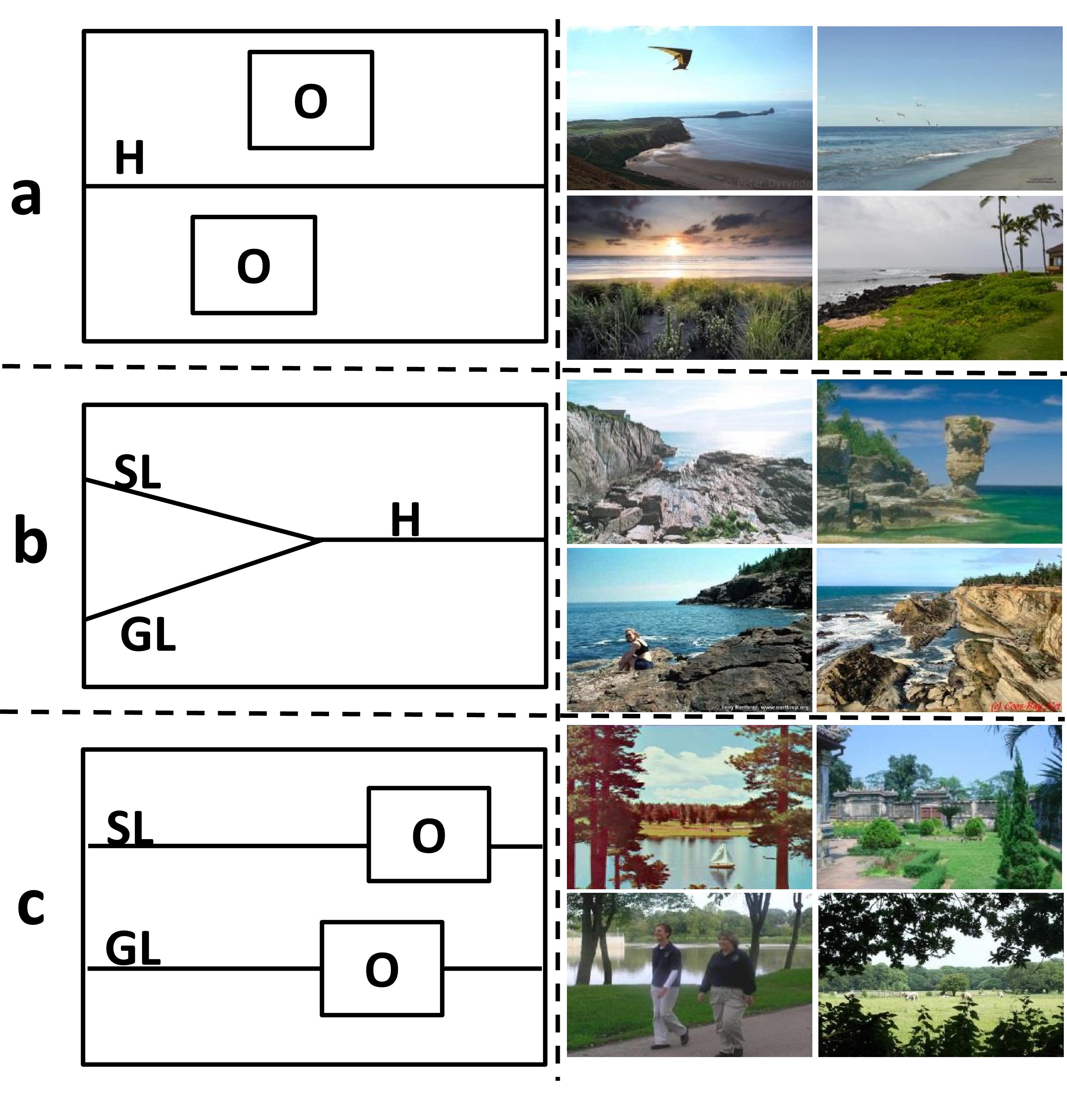}
\end{center}
\caption{A proposed framework for geometric layout reasoning: (a) the
simplest case, (b) the background scene only, and (c) a general scene,
where SL, GL, H, O denote the sky line, the ground line, the horizon and
the occluder, respectively.} \label{fig:LR2}
\end{figure}

The 7D GAV can characterize a wide range of scene images. In this
section, we focus on layout reasoning using the extracted global
attribute vector. We first analyze the outdoor scenes from the simplest
setting to the general setting as shown in \figurename{\ref{fig:LR2}}.
Then, we propose a Conditional Random Field (CRF) minimization framework
for outdoor layout reasoning. 

The simplest case of our interest is a scene with a horizon as shown in
\figurename{\ref{fig:LR2}(a)}. This leads to sky and support two regions
and the support could be ocean or ground. The existence of occluders
makes this case slightly more complicated. The sky may have two
occluders - solid (e.g., balloon, bird, airplane, kite, parashoot) and
porous (e.g., tree leaves near the camera). The ocean/ground may also
have occluders - solid (e.g, boat in the ocean, fence/wall near the
camera) and porous (e.g. bushes near the camera). Although there are
rich scene varieties, our layout reasoning framework can remove segments
labeled with planar center/left/right in the simplest case confidently. 

The horizon will be split into individual sky and ground lines if there
exist large planar surfaces that completely or partially block the
horizon as shown in \figurename{\ref{fig:LR2}(b)}. It is a background
scene only, if there is no occluder. The large planar surfaces can be
buildings in urban scenes and mountains in natural scenes.  Generally
speaking, any region between the sky and ground lines is a strong
candidate for planar surfaces. We can further classify them into planar
left/right/center based on vanishing lines or the outer contour of the
surface region.  A general scene may have all kinds of solid/porous
occluders (e.g.  people, car, trees, etc.) in front of a background
scene as shown in \figurename{\ref{fig:LR2}(c)}. The occluders may block
the ground line, the planar surface and other solid/porous objects to
make layout inference very challenging. In the following, we exploit
initial pixel labeling and all attributes and formulate layout reasoning
as a Conditional Random Field (CRF) \cite{fulkerson2009class},
\cite{song2015efficient} optimization problem. 

{\bf Problem Formulation:} We model each image as a graph denoted by
$G(S,E)$, where each node is the super-pixel segment and a member of $S$
and two adjacent vertices are connected by an edge which is a member of
$E$. Our goal is to minimize the following energy function given the
adjacency graph $G(S,E)$ and a weight $\lambda$:
\begin{align}\label{eq:energyfuc}
E(\mathbf{L}) &= \sum_{s_i\in S}\Psi(l_i|s_i)
+\lambda \sum_{(s_i,s_j)\in E}\Phi(l_i,l_j|s_i,s_j),
\end{align}
where $S$ is the segmentation of consideration, $l_{i}$ is the class
label assigned to the $i^{th}$ segment $s_{i}$, $\Psi$ is the unary
potential function, $\Phi$ is the pairwise potential function, and
$\lambda$ is a tradeoff between the unary potential and the pairwise
potential. Empirically, we set $\lambda=0.1$. 

We classify the seven proposed global attributes into two major groups.
The first group consists of the horizon, the vertical line, the solid
and the porous.  Their probability maps are determined by the appearance
properties of segments rather than the context relationship among
adjacent segments. It is proper to adopt the unary potential as their
probability outputs. The second group consists of the sky/ground line,
the vanishing line, the left or right-oriented surfaces, which describe
the relationship among adjacent segments. Thus, we consider the pairwise
potential of these attributes.  Specifically, we have the following
definitions of the unary potential and the pairwise potential. 

We define the unary potential for each segment $s_{i}$ as 
\begin{align}\label{eq:unary}
\Psi(l_i|s_i) = -\log(P(l_i|s_i)),
\end{align}
where
\begin{equation}      
P(l_i|s_i) = {\textbf{w}^\intercal} \left(  
\begin{array}{c}   
P_{\text{initial}}(l_i|s_i) \\
P_{\text{porous}}(l_i|s_i)\\
P_{\text{solid}}(l_i|s_i)\\
P_{\text{horizon}}(l_i|s_i)\\
P_{\text{verticall}}(l_i|s_i)\\
\end{array}
\right)                
\end{equation}
is the probability that segment $s_i$ is predicted with label $l_i$.  In
words, probability $P(l_i|s_i)$ is the weighted average of five
components: 1) initial label probability $P_{\text{initial}}(l_i|s_i)$ which
comes from the label probability from our 3 class classifier and
vertical class classifier from \cite{hoiem2005geometric}; 2) porous
detector $P_{\text{porous}}(l_i|s_i)$; 3) solid detector $P_{\text{solid}}(l_i|s_i)$;
4) horizon line inference $P_{\text{horizon}}(l_i|s_i)$; and 5) vertical line
inference $P_{\text{verticall}}(l_i|s_i)$. The weight vector $\textbf{w}$ is a
5D vector that determines the contributions of the five components in
the unary potential. 

We define the pairwise potential for pairwise segments $s_{i}$ and
$s_{j}$ as
\begin{align} \label{eq:pairwise}
\Phi(l_i,l_j|s_i,s_j) =(\Phi_{s}+\Phi_{v}+\Phi_{p})[l_{i} \neq l_{j}],
\end{align}
where $[\cdot]$ is a zero-one indicator function and

\begin{flalign}
&\Phi_s=-\log(P_{\text{sky/ground line}_\text{b}}(s_i, s_j)), \label{eq:pairwise_s} \\
&\Phi_v=-\log(\|P_{\text{vanishing line}}(s_i, l_i)-P_{\text{vanishing line}}(s_j, l_j)\|), \label{eq:pairwise_v}\\
&\Phi_p=-\log(\|P_{\text{planar surfance}}(s_i, l_i)-P_{\text{planar surface}}(s_j, l_j)\|) \label{eq:pairwise_p}
\end{flalign}
where $P_{\text{sky/ground line}_\text{b}}(s_i, s_j)$, $P_{\text{vanishing line}}(s_i, l_i)$, and $P_{\text{planar surfance}}(s_i, l_i)$ denote averaged probability along the shared boundary of $s_{i}$ and $s_{j}$, probability of $s_{i}$ is labeled as $l_{i}$ from vanishing line information, and probability of $s_{i}$ is labeled as $l_{i}$ from planar surface information, respectively, and $\|\cdot\|$ is the norm of the probability difference between
segments $s_{i}$ and $s_{j}$.
 
In Eq. \ref{eq:pairwise_s}, when there is neither sky nor ground line across $s_{i}$ and $s_{j}$,
$P_{\text{sky/ground line}_\text{b}}(s_i, s_j)$ should be small. However, $\Phi_{s}$
would be large if $s_{i}$ and $s_{j}$ are with different labels.  To
minimize potential function in Eq. \ref{eq:energyfuc}, $\Phi_{s}$ forces
$s_{i}$ and $s_{j}$ to be with the same label. This explains how the
ground line is removed in the example given in Fig.
\ref{fig:sky_ground_existence}.  Furthermore, $\Phi_{v}$ and $\Phi_{p}$
are the pairwise potentials computed from the probability map of
vanishing lines and the probability map of planar surface orientation as
shown in Eqs. \ref{eq:pairwise_v} and \ref{eq:pairwise_p}, respectively.
A smoothness constraint is imposed by $\Phi_{v}$ (or $\Phi_{p}$).
$\Phi_{v}$ (or $\Phi_{p}$) penalizes more if $s_{i}$ and $s_{j}$ have
different labels yet their vanishing line (or planar surface
orientation) difference at location $s_{i}$ and $s_{j}$ is small.  This
explains how the surface orientation smoothness can be achieved in the
examples given in \figurename{\ref{fig:surface}}. 

We first learn the parameters, \textbf{w} and $\lambda$, from the CRF
model by maximizing their conditional likelihood through
cross-validation on the training data. Then, the multi-label graph cut
optimization method \cite{fulkerson2009class} is used to minimize the
energy function in Eq. \ref{eq:energyfuc}. 

\section{Experimental Results}\label{sec:result}

In contrast to the object detection problem, very few datasets are
available for geometric layout algorithms evaluation. The dataset in
\cite{RefWorks:393} is the largest benchmarking dataset in the geometric
layout research community. There are larger geometric layout datasets
but for indoor images which doesn't fit our problem
\cite{choi2013understanding}, \cite{hedau2009recovering}. The dataset in
\cite{RefWorks:393} consists of 300 images of outdoor scenes with
labeled ground truth. The first 50 images are used for training the
surface segmentation algorithm as done in previous work
\cite{RefWorks:393}, \cite{RefWorks:389}. The remaining 250 images are
used for evaluation. We follow the same procedure in our experiments. To
test our generalized global attributes and reasoning system, we use the
same 250 testing images in our experiment. 

It is not a straightforward task to label the ground truth for outdoor
scene images in this dataset. It demands subjects to reason the
functionalities of different surfaces in the 3D real world (rather than
recognizing objects in images only), e.g., where is the occlusion
boundary? what is the surface orientation? whether there is depth
difference? etc. To get a consistent ground truth, subjects should be
very well trained. Note that we do not agree with the labeled ground
truth of several images, and will report them in the supplemental
material. In this section, we will present both qualitative and
quantitative evaluation results and analyze several poorly labeled
images.

\begin{figure*}[htbp]
\begin{center}
\includegraphics[width=0.8\textwidth]{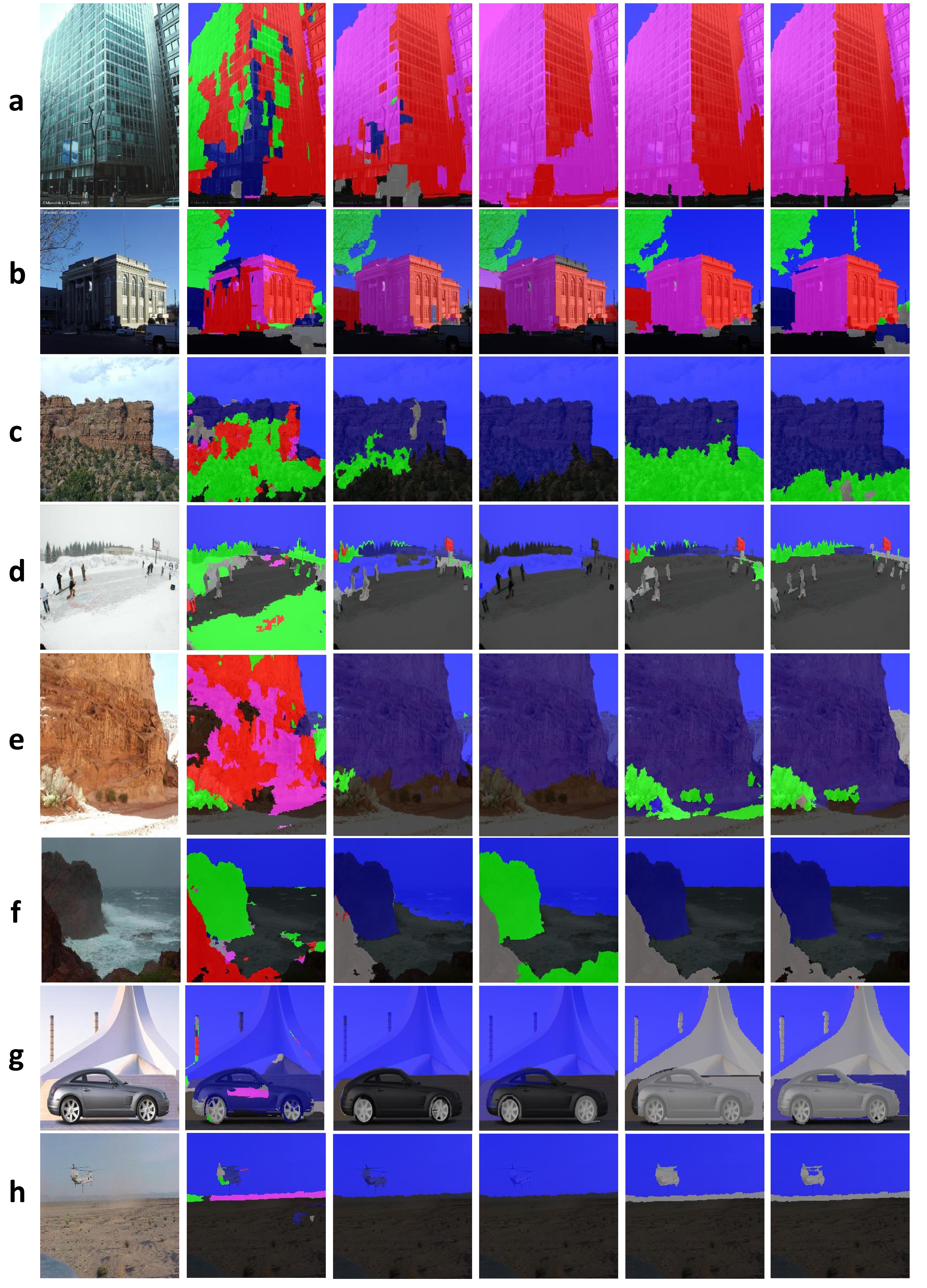}
\caption{Qualitative comparisons of three geometric
layout algorithms (from left to right): the original Image,
CNN-method\cite{badrinarayanan2015segnet}, Hoiem et al.
\cite{RefWorks:393}, Gupta et al. \cite{RefWorks:389}, the GAL and the
ground truth.  The surface layout color codes are magenta (planar
left), dark blue (planar center), red (planar right), green (non-planar
porous), gray (non-planar solid), light blue (sky), black (support).} 
\label{fig:result_comparsion} \end{center}
\end{figure*}

\textbf{Qualitative Analysis.} \figurename{\ref{fig:result_comparsion}}
shows eight representative images (the first column) with their labeled
ground truths (the last column). We show results obtained by CNN method
\cite{badrinarayanan2015segnet}(the second column), Hoiem et al.
\cite{RefWorks:393} (the third column), Gupta et al. \cite{RefWorks:389}
(the fourth column), the proposed GAL (the fifth column).  To apply the
CNN to this problem, we retrained the CNN use the training data with
initialized weights from the VGG network
\cite{badrinarayanan2015segnet}. We call the methods proposed in Hoiem
et al. \cite{RefWorks:393} and Gupta et al. \cite{RefWorks:389} the
H-method and the G-method, respectively, in the following discussion for
convenience. 
 
In \figurename{\ref{fig:result_comparsion} (a)}, the building is tilted
and its facade contains multiple surfaces with different orientations.
Because of the similarity of the local visual pattern (color and
texture), the CNN method got confused between surface and porous. The
H-method is confused by upper and lower parts in the planar-left facade,
and assigns them to planar left and right two opposite oriented
surfaces. This error can be corrected by the global attribute (ie
building's vertical lines). The G-method loses the detail of surface
orientation in the upper left corner so that the sky and the building
surface are combined into one single planar left surface. It also loses
the ground support region. These two errors can be corrected by sky-line
and ground-line detection. GAL infers rich semantic information about
the image from the 7D GAV. Both the sky and the ground exist in the
image. Furthermore, it is a building facade due to the existence of
parallel vertical lines. It has two strong left and right vanishing
points so that it has two oriented surfaces - one planar left and the
other planar right. There is a problem remaining in the intersection
region of multiple buildings in the right part of the image. Even humans
may not have an agreement on the number of buildings in that area (two,
three, or four?) The ground truth also looks strange. It demands more
unambiguous global information for scene layout inference. 

For \figurename{\ref{fig:result_comparsion} (b)}, the ground truth has
two problems. One is in the left part of the image. It appears to be
more reasonable to label the building facade to ``planar left" rather
than ``planar center". Another is in the truck car region. The whole
truck should have the same label; namely, gray (non-planar solid). This
example demonstrates the challenging nature of the geometric layout
problem. Even human can make mistakes easily. Other observed
human-labeling mistakes in ground truth are reported in the supplemental
file. 

\figurename{\ref{fig:result_comparsion} (c)-(f)} show the benefit of
getting an accurate sky line and/or an accurate ground line. One
powerful technique to find the sky line and the ground line is to use
defocus estimation. Once these two lines are detected, the inference
becomes easier so that GAL yields better results.

Being constrained by the limited model hypothesis, the G-method chooses
to have the ground (rather than the porous) to support the physical
world. However, this model does not apply to the image in (c), images
with bushes in the bottom as an occluder, etc. Also, the ground truth in
(c) is not consistent in the lower region. The result obtained by GAL
appears to be more reasonable.

\figurename{\ref{fig:result_comparsion} (d)} was discussed earlier. Both
the H-method and the G-method labeled part of the ground as the sky by
mistake due to color similarity. GAL can fix this problem using horizon
line detection. All three methods work well for \figurename{\ref{fig:result_comparsion}} (e). GAL performs slightly better due to
the use of the horizon detection and label refinement stage. The right
part of the image is too blurred to be of interest. GAL outperforms the
CNN-method, the H-method and the G-method due to an accurate horizon
detection in \figurename{\ref{fig:result_comparsion} (f)}. \figurename{\ref{fig:result_comparsion} (g)} demonstrates the power of the sky line
detection and object (car) detector. \figurename{\ref{fig:result_comparsion} (h)} demonstrates the importance of an
accurate sky line detection. 

\begin{figure}[htbp]
\begin{center}
\includegraphics[width=1\linewidth]{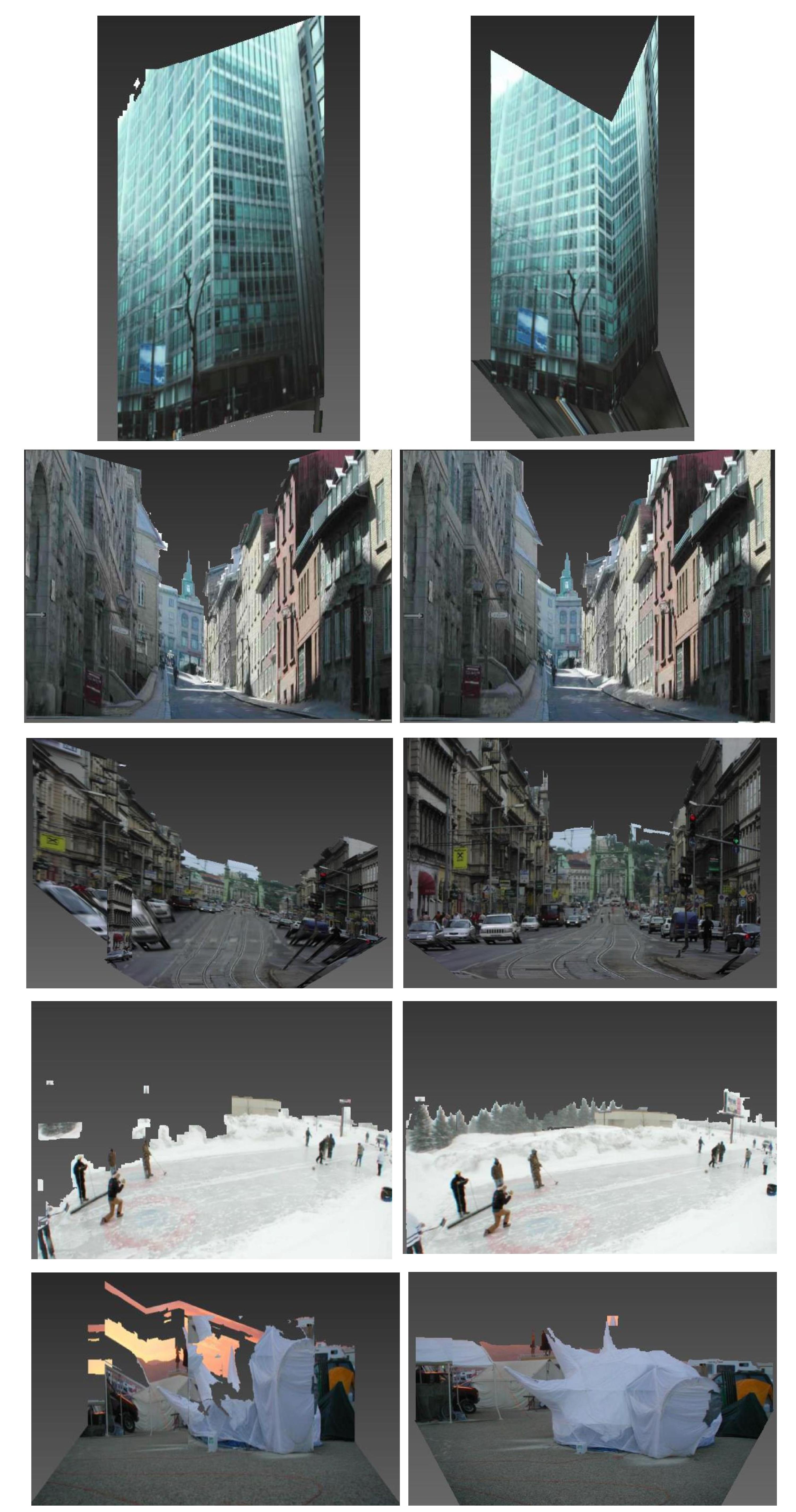}	
\caption{Comparison of 3D rendered views based on geometric labels from
the H-method (left) and GAL (right).}\label{fig:pop_up}
\end{center} 
\end{figure}

We also compare the ``pop-up" view of the H-method and GAL in
\figurename{\ref{fig:pop_up}} using a 3D view rendering technique given
in \cite{RefWorks:410}, \cite{RefWorks:392}. It is clear that GAL offers a more meaningful
result. 

Based on the above discussion, we can draw several concluding points.
First, there are not enough examples for the CNN to learn the visual
pattern. The CNN is a data driven approach that extract visual patterns
from training data.  For the geometric labeling problem, labels cannot
be fully determined by visual patterns. For example, the left and right
surface of a building have similar color and texture visual pattern
locally, but their orientations are different.  In contrast, the
orientation can be captured by the global information such as shapes and
lines. With very little training data, CNN's power is somehow limited.
Second, the local-patch-based machine learning algorithm (e.g., the
H-method) lacks the global information for 3D layout reasoning. It is
difficult to develop complete outdoor scene models to cover a wide
diversity of outdoor images (e.g., the G-method).  GAL attempts to find
key global attributes from each outdoor image. As long as these global
attributes can be determined accurately, we can combine the local
properties (in form of initial labels) and global attributes for layout
reasoning and label refinement. Third, the proposed methodology can
handle a wider range of image content. As illustrated above, it can
provide more satisfactory results to quite a few difficult cases that
confuses prior art such as the H-method and the G-method. 

\textbf{Quantitative Analysis.} For quantitative performance evaluation,
we use the metric used in \cite{RefWorks:392,RefWorks:389} by computing
the percentage of overlapping of labeled pixel and ground truth pixel
(called the pixel-wise labeling accuracy). We use the original ground
truth given in \cite{hoiem2005geometric} for fair comparison between all
experimental results reported below (with only one exception which will
be clearly stated). 

\begin{figure*}[t]
\begin{center}
\includegraphics[width=0.8\textwidth]{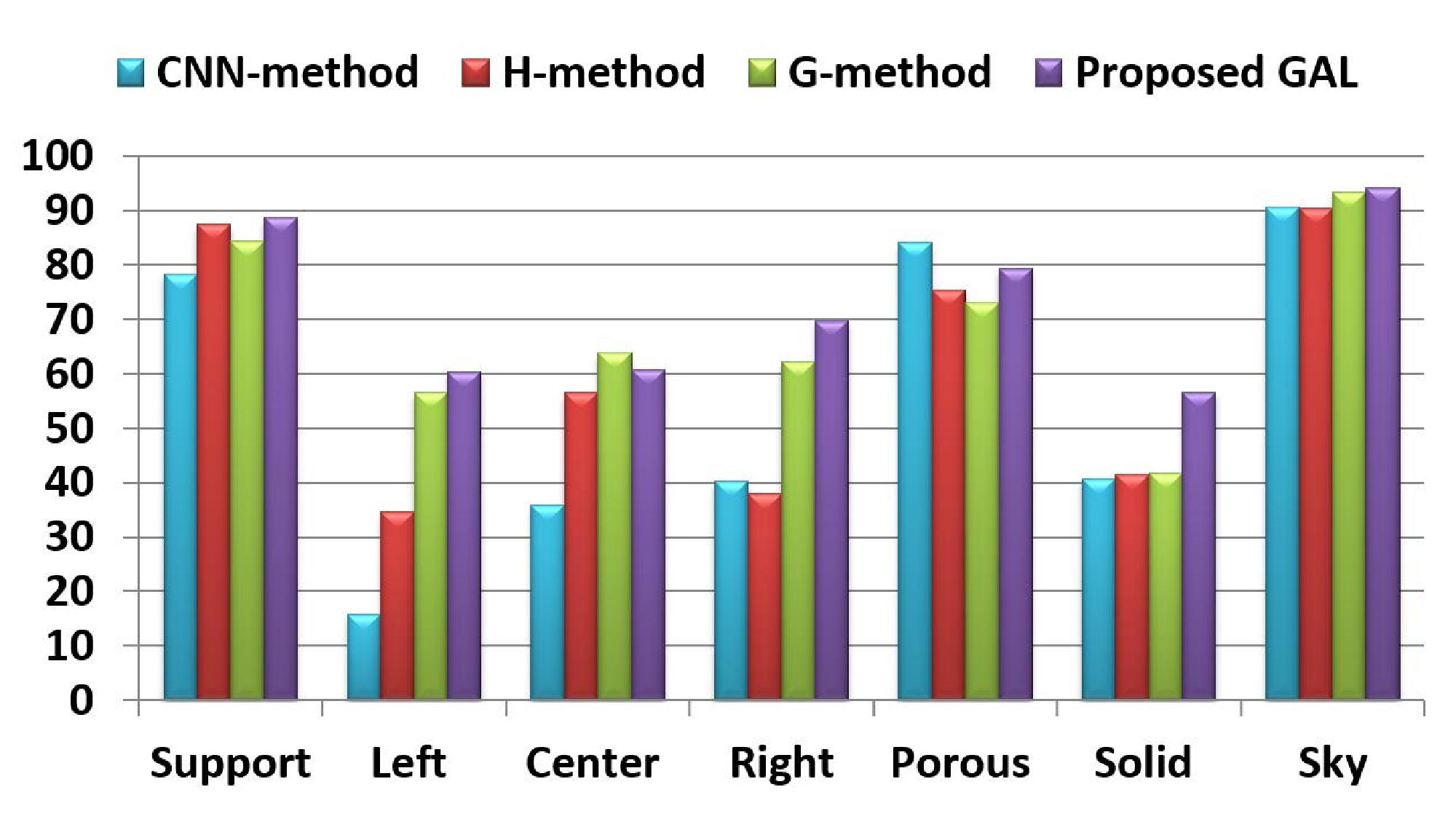}		
\caption{Comparison of labeling accuracy between the CNN methods, the
H-method, the G-method and GAL with respect to seven individual labels.}
\label{fig:performance}
\end{center}
\end{figure*}	

We first compare the labeling accuracy of the CNN-method, the H-method,
the G-method and GAL for seven geometric classes individually in
\figurename{\ref{fig:performance}}. We see clearly that GAL outperforms
the H-method in all seven classes by a significant margin. The gain
ranges from 1.27\% (support) to more than 25\% (planar left and right).
GAL also outperforms the G-method in 6 classes.  The gain ranges from
0.81\% (sky) to 14.9\% (solid). The G-method and GAL have comparable
performance with respect to the ``planar center" class. All methods do
well in sky and ground labeling. The solid class is the most challenging
one among all seven classes since it has few common geometric structures
to exploit. 

The complete performance benchmarking (in terms of pixel-wise labeling
accuracy) of five different methods are shown in Table \ref{table_1}.
The L-method in \cite{RefWorks:402} and the P-method in
\cite{RefWorks:385} were dedicated to building facade labeling and their
performance were only evaluated on subsets of the full dataset in
\cite{hoiem2005geometric}. The L-method in \cite{RefWorks:402} use the
subset which contains 100 images images where the ground truth of both
occlusion boundaries and surface orientation are provided
\cite{RefWorks:393}. P-method in \cite{RefWorks:385} use the subset
which contains 55 building images. 

We use B1, B2, F and F/R to denote the subset used in
\cite{RefWorks:402}, the subset used in \cite{RefWorks:385}, the full
dataset and the full dataset with relabeled ground truth in the table,
respectively.  The two numbers within the parenthesis (7 and 5) denote
results for all ``seven" classes and for the ``five" vertical classes
(i.e., excluding sky and support), respectively. For the F(7) column in
Table \ref{table_1}, we compare the performance for all seven classes in
the full set. Accurate labeling for this dataset is actually very
challenging as pointed out in \cite{lazebnik2009empirical},
\cite{ramalingam2008exact}. This is also evidenced by the slow
performance improvement over the last seven years - a gain of 2.35\%
from the H-method to the G-method. GAL offers another gain of 4.95\%
over the G-method \cite{RefWorks:389}, which is significant. For the
B1(5) column, we compare the performance for five classes in the
building subset which contains 100 images. GAL outperforms the
CNN-method and the L-method by 21.24\% and 3.39\%, respectively. For the
B2(7) column, we compare the performance against the building subset
which contains 55 images. GAL outperforms the CNN-method, the H-method,
the G-method and the P-method by 20.4\%, 8.8\%, 8.08\% and 6.85\%,
respectively. For the F(5) column, we compare the performance for five
classes in the full set. GAL outperforms the CNN-method, the H-method,
the G-method by 19.64\%, 7.14\% and 2.22\%.  Finally, for the F/R(7)
column, we show the labeling accuracy of GAL against the modified ground
truth.  The labeling accuracy can go up to 80.26\%. 
 
\begin{table}[!ht]
\caption{Comparison of the averaged labeling accuracy (\%) of six
methods with respect to the building subset (B), the full set (F)
and the full set with relabeled ground truth (F/R), where 7 and 5
mean all seven classes and the five classes belonging to the vertical
category, respectively.Results are updated}\label{table_1}
\begin{center}\small
\begin{tabular}{|l||c|c|c|c|c|}
\cline{2-6}
\multicolumn{1}{l|}{}        & \multicolumn{5}{c|}{Dataset (Class No.)}\\ \cline{2-6}
\multicolumn{1}{l|}{}        & B1(5)  & B2(7)  & F(5)  & F(7)  & \!F/R(7)\! \\ \hline
CNN-method \cite{badrinarayanan2015segnet} &  58.33  & 61.27 & 56.33  & 68.21  & N/A \\ \hline
H-method \cite{RefWorks:393} & N/A   & 72.87 & 68.80 & 72.41 & N/A    \\ \hline
G-method \cite{RefWorks:389} & N/A   & 73.59 & 73.72 & 74.76 & N/A    \\ \hline
L-method \cite{RefWorks:402} & 76.34 & N/A   & N/A   & N/A   & N/A    \\ \hline
P-method \cite{RefWorks:385} & N/A   & 74.82 & N/A   & N/A   & N/A    \\ \hline 
Proposed GAL                 & 79.73 & 81.67 & 75.94 & 79.71 & 80.26  \\ \hline
\end{tabular}
\end{center}\normalsize
\end{table}

To further analyze the performance gain of proposed GAL system, we add
the global attributes one by one to our proposed GAL system and analyze performance gain from individual global attribute, which is
shown in Table.2. The initial result for 7 class tested in the full set is 74.88\% which is obtained by our 3 class labeling result and the labels in vertical region from \cite{hoiem2005geometric}. When compute the performance gain from global attribute horizon, $P_{\text{horizon}}$ is first computed and $P_{\text{horizon}}$ is included in the unary term in CRF model. As there is no global attributes contribute to pairwise term, we simply apply the same pairwise edge potentials defined in \cite{fulkerson2009class}. As both attribute planar surface and attribute vanishing line contribute to surface orientation accuracy improvement, we include both of them to see their performance gain together.    
The table shows that all global attributes contributes to the
performance gain significantly. 

\begin{table}[!ht]
\caption{The performance gain from each individual attribute.}\label{table_2}
\begin{center}\small
\begin{tabular}{|l||c|c|c|c|c|c|c|}  \hline 
     & porous   & solid        &  horizon    \\ \hline 
Performance Gain &  +0.12\%   &  + 0.80\%           &  + 1.08\% \\ \hline\hline
     &  vertical  &  sky/ground     & vanishing line  \\  
      &  line         &        line  & planar surface  \\ \hline                     
Performance Gain &  + 1.00\% &  + 0.48\%           & + 0.95\% \\ \hline
\end{tabular}
\end{center}\normalsize
\end{table}

\begin{figure}
\begin{center}
\includegraphics[width=1.0\linewidth]{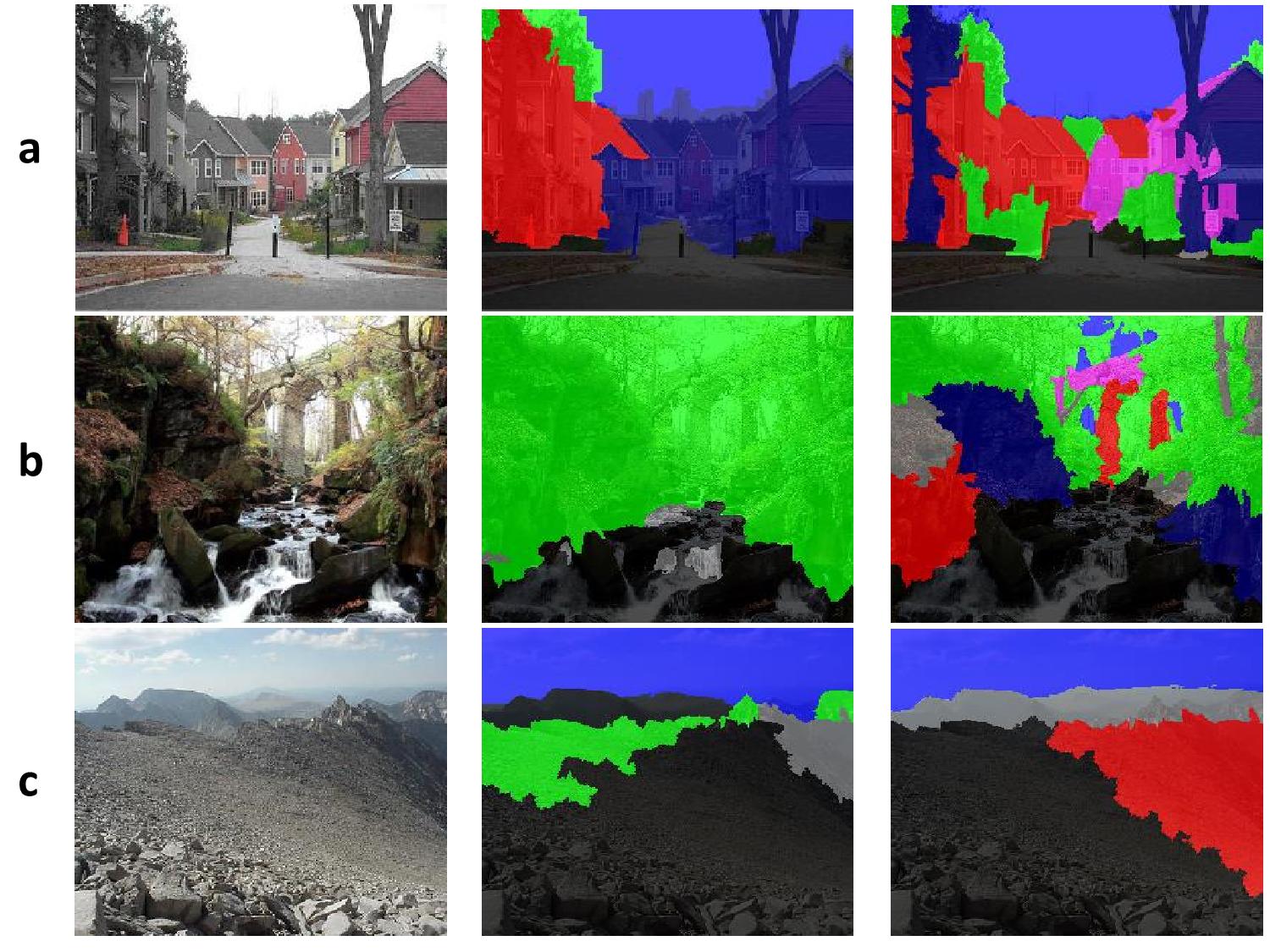}
\end{center}
\caption{Error analysis of the proposed GAL system with three exemplary
images (one example per row and from left to right): the original image,
the labeled result of GAL and the ground truth.}\label{fig:error}
\end{figure}

\textbf{Error Analysis.} Three exemplary images that have large labeling
errors are shown in \figurename{\ref{fig:error}}.
\figurename{\ref{fig:error} (a)} is difficult due to the complexity in
the middle region of multiple houses. We feel that the labeled ``planar
center" result by GAL for this region is still reasonable although it is
not as refined as that offered by the ground truth.
\figurename{\ref{fig:error} (b)} is one of the most challenging scenes in
the dataset since even humans may have disagreement.  The current ground
truth appears to be too complicated to be useful. GAL can find the
ground line but not the sky line due to the tree texture in the top part
of the image. GAL made a labeling mistake in porous in
\figurename{\ref{fig:error} (c)} due to dominant texture in the
corresponding region. We need a better porous detector to fix this
problem. The global attributes will not be able to help much. There is a
planar right region in the ground truth. Although this label is accurate
to human eyes by looking at the local surface, the transition from
support (or ground) to planar right is not natural. It could be an
alternative to treat the whole region in the bottom part as support.  As
shown in this image, it is extremely difficult to find an exhaustive set
of scene models to fit all different situations. 

\section{Conclusion}\label{sec:conclusion}

A novel GAL geometric layout labeling system for outdoor scene images
was proposed in this work. GAL exploits both local and global attributes
to achieve higher accuracy and it offers a major advancement in solving
this challenging problem. Its performance was analyzed both
qualitatively and quantitatively. Besides, several error cases were
studied to reveal the limitations of the proposed GAL system. 

Clearly, there are still many interesting problems remaining, including
the development of better global attributes extraction tools, its
integration with CNN detectors, and the design of more powerful and more
general inference rules. Furthermore, due to major differences between
indoor and outdoor scene images, the key global attributes of indoor
scene images will be different from those of outdoor scene images. So,
more research needs to be done to develop a GAL system for indoor
scenes. Future research is planned to determine important global
attributes of indoor scene images and to study their geometric layout. 
 
\ifCLASSOPTIONcaptionsoff
 \newpage
\fi
 
\bibliographystyle{IEEEtran}
 
\bibliography{TIP_v7}

\end{document}